\documentclass[10pt,,compsoc]{IEEEtran}

\usepackage{amsmath}
\usepackage{mathtools}
\usepackage{graphics,epsfig}
\usepackage{xspace}
\usepackage{algorithm}
\usepackage{algorithmic}
\usepackage{amssymb}
\usepackage{subfigure}
\usepackage{multirow}
\usepackage{multicol}
\usepackage{lineno,hyperref}

\begin{document}

\title{Boost K-Means}
\author{Wan-Lei~Zhao,
        Cheng-Hao~Deng,
        and~Chong-Wah~Ngo,~\IEEEmembership{Senior~Memeber,~IEEE}
\IEEEcompsocitemizethanks{\IEEEcompsocthanksitem Fujian Key Laboratory of Sensing and Computing for Smart City, and the School of Information Science and Engineering, Xiamen University, Xiamen, 361005, P. R. China.\protect\\
E-mail: wlzhao@xmu.edu.cn
\IEEEcompsocthanksitem Department of Computer Science, City University of Hong Kong.\protect\\
E-mail: cscwngo@gapps.cityu.edu.hk}}

\IEEEtitleabstractindextext{
\begin{abstract}
Due to its simplicity and versatility, k-means remains popular since it was proposed three decades ago. The performance of k-means has been enhanced from different perspectives over the years. Unfortunately, a good trade-off between quality and efficiency is hardly reached. In this paper, a novel k-means variant is presented. Different from most of k-means variants, the clustering procedure is driven by an explicit objective function, which is feasible for the whole $\textit{l}_2$-space. The classic egg-chicken loop in k-means has been simplified to a pure stochastic optimization procedure. The procedure of k-means becomes simpler and converges to a considerably better local optima. The effectiveness of this new variant has been studied extensively in different contexts, such as document clustering, nearest neighbor search and image clustering. Superior performance is observed across different scenarios.
\end{abstract}

\begin{IEEEkeywords}
clustering, k-means, incremental optimization
\end{IEEEkeywords}}

\maketitle

\section{Introduction}
\label{sec:intro}
Clustering problems arise from variety of applications, such as documents/web pages clustering~\cite{ml04:zhao}, pattern recognition, image linking~\cite{ikmn15}, image segmentation~\cite{normcut}, data compression via vector quantization~\cite{SiZ03} and nearest neighbor search~\cite{JDS11,pami14:flann,cvpr14:artem}. In the last three decades, various clustering algorithms have been proposed. Among these algorithms, k-means~\cite{km82} remains a popular choice for its simplicity, efficiency and moderate but stable performance across different problems. It was known as one of top ten most popular algorithms in data mining~\cite{top10}. On one hand, k-means has been widely adopted in different applications. On the other hand, continuous efforts have been devoted to enhance the performance k-means as well.

Despite its popularity, it actually suffers from several latent issues. Although the time complexity is linear to data size, traditional k-means is still not sufficiently efficient to handle the web-scale data. In some specific scenarios, the running time of k-means could be even exponential in the worst case~\cite{ailon09,vattani11}. Moreover, k-means usually only converges to local optima. As a consequence, recent research has been working on either improving its clustering quality~\cite{kpp07,nips11:wong} or efficiency~\cite{dmount02,icml03:elkan,mnkm10,nips11:wong,kpp12,wsdm14}. K-means has been also tailored to perform web-scale image clustering~\cite{ikmn15,photohash15}.

There are in general three steps involved in the clustering procedure. Namely, 1. initialize \textit{k} cluster centroids; 2. assign each sample to its closest centroid; 3. recompute cluster centroids with assignments produced in Step 2 and go back to Step 2 until convergence. This is known as \textit{Lloyd} iteration procedure~\cite{km82}. The iteration repeats Step 2 and Step 3 until the centroids do not change between two consecutive rounds. Given $C_{1{\cdots}k} \in R^d$ are cluster centroids, $\{x_i \in R^d\}_{i=1{\cdots}n}$ are samples to be clustered, above procedure essentially minimizes the following objective function:
\begin{equation}
        \mbox{min }\sum_{q(x_i)=r}{\parallel C_r -x_i \parallel^2}.
        \label{eqn:tkm}
\end{equation}
In Eqn.~\ref{eqn:tkm}, function \textit{q}($\cdot$) returns the closest centroid for sample $x_i$. Unfortunately, searching an optimal solution for the above objective function is NP-hard. In general k-means only converges to local minimum~\cite{bottou95}. The reason that k-means maintains its popularity is mainly due to its linear complexity in terms of the number of samples to be clustered. The complexity is $O(t{\cdot}k{\cdot}n{\cdot}d)$, given $t$ as the number of iterations to converge. Compared with other well-known clustering algorithms such as DBSCAN~\cite{dbscan} and Mean shift~\cite{meanshift02}, this complexity is considerably low. However, the efficiency of traditional k-means cannot cope with the massive growth of data in Internet. In particular, in the case that the size of data (\textit{m}), the number of clusters (\textit{k}) and the dimension (\textit{d}) are all very large, k-means becomes unbearably slow. The existing efforts~\cite{mnkm10,wsdm14} in enhancing the scalability of k-means for web-scale tasks often come with price of lower clustering quality.
On the other hand, k-means++ proposed in~\cite{kpp07,kpp12} focuses on enhancing the clustering quality by a careful design of the initialization procedure. However, k-means is slow down as a few rounds of scanning over the dataset is still necessary in the initialization.

In this paper, a novel variant of k-means is proposed, which aims to make a better trade-off between clustering quality and efficiency. Inspired by the work in~\cite{ml04:zhao}, a novel objective function is derived from Eqn.~\ref{eqn:tkm}. With the development of this objective function, the traditional k-means iteration procedure has been revised to a simpler form, in which the costly initial assignment becomes unnecessary. In addition, driven by the objective function, sample is moved from one cluster to another cluster when we find this movement leads to higher objective function score, which is known as incremental clustering~\cite{ml04:zhao,pttnc01}. These modifications lead to several advantages.
\begin{itemize}
        \item {K-means clustering without initial assignment results in better quality as well as higher speed efficiency.}
        \item {K-means iteration driven by an explicit objective function converges to considerably lower clustering distortion in faster pace.}
        \item {Different from traditional k-means, it is not necessary to assign a sample to its closest centroid in each iteration, which also leads to higher speed.}
\end{itemize}

In addition, when clustering in hierarchical bisecting fashion, the proposed method achieves the highest scalability among all top-down hierarchical clustering methods. Extensive experiments are conducted to contrast the performance of proposed method with k-means and its variants including tasks document clustering~\cite{ml04:zhao}, nearest neighbor search (NNS) with product quantization~\cite{SiZ03} and image clustering.

The remainder of this paper is organized as follows. The reviews about representative works on improving the performance of traditional k-means are presented in Section~\ref{sec:relat}. In Section~\ref{sec:func}, the clustering objective functions are derived based on Eqn.~\ref{eqn:tkm}. Based on the objective function, Section~\ref{sec:alg} presents the clustering method. Extensive experiement studies over proposed clustering method are presented in Section~\ref{sec:exp}. Section~\ref{sec:conl} concludes the paper.
\section{Related Works}
\label{sec:relat}
Clustering is a process of partitioning a set of samples into a number of groups without any supervised training. Due to its versatility in different contexts, it has been studied in the last three decades~\cite{jain99}. As the introduction of Web 2.0, millions of data in Internet has been generated on a daily basis. Clustering becomes one of the basic tools to process such big volume of data. As a consequence, traditional clustering methods have been shed with new light. People are searching for clustering methods that are scalable~\cite{mnkm10,kpp12,wsdm14} to web-scale data. In general, boosting the performance of traditional k-means becomes the major trend due to its simplicity and relative higher efficiency over other clustering methods. 

In general, there are two major ways to enhance the performance of k-means. For the first kind, the aim is to improve the clustering quality. One of the important work comes from Ostrovsky et al.~\cite{kpp07,kpp12}. The motivation is based on the observation that k-means converges to a better local optima if the initial cluster centroids are carefully selected. According to~\cite{kpp07}, k-means iteration also converges faster due to the careful selection on the initial cluster centroids. However, in order to adapt the initial centroids to the data distribution, \textit{k} rounds of scanning over the data are necessary. Although the number of scanning rounds has been reduced to a few in~\cite{kpp12}, the extra computational cost is still inevitable.

In each k-means iteration, the processing bottleneck is the operation of assigning each sample to its closest centroid. The iteration becomes unbearably slow when both the size and the dimension of the data are very large. Noticed that this is a nearest neighbor search problem, Kanungo et al.~\cite{dmount02} proposed to index dataset in a KD Tree~\cite{kdtree75} to speed-up the sample-to-centroid nearest neighbor search. However, this is only feasible when the dimension of data is in few tens. Similar scheme has been adopted by Dan et al.~\cite{pelleg99}. Unfortunately, due to the curse of dimensionality, this method becomes ineffective when the dimension of data grows to a few hundreds. A recent work~\cite{wsdm14} takes similar way to speed-up the nearest neighbor search by indexing dataset with inverted file structure. During the iteration, each centroid is queried against all the indexed data. Attributing to the efficiency of inverted file structure, one to two orders of magnitude speed-up is observed. However, inverted file indexing structure is only effective for sparse vectors.

Alternatively, the scalability issue of k-means is addressed by subsampling over the dataset during k-means iteration. Namely, methods in~\cite{mnkm10,icdm04} only pick a small portion of the whole dataset to update the cluster centroids each time. For the sake of speed efficiency, the number of iterations is empirically set to small value. It is therefore possible that the clustering terminates without a single pass over the whole dataset, which leads to higher speed but also higher clustering distortion. Even though, when coping with high dimensional data in big size, the speed-up achieved by these methods is still limited.

Apart from above methods, there is another easy way to reduce the number of comparisons between samples and centroids, namely performing clustering in a top-down hierarchical manner~\cite{ml04:zhao,jain88, kddzhao05}. Specifically, the clustering solution is obtained via a sequence of repeated bisections. The clustering complexity of k-means is reduced from $O(t{\cdot}k{\cdot}n{\cdot}d)$ to $O(t{\cdot}log(k){\cdot}n{\cdot}d)$. This is particularly significant when \textit{n}, \textit{d} and \textit{k} are all very large. In addition to that, another interesting idea from~\cite{ml04:zhao, kddzhao05} is that cluster centroids are updated incrementally~\cite{ml04:zhao,pttnc01}. Moreover, the update process is explicitly driven by an objective function (called as criterion function in~\cite{ml04:zhao,kddzhao05}). Unfortunately, objective functions proposed in~\cite{ml04:zhao,jain88, kddzhao05} are based on the assumption that input data are in unit length. The clustering method is solely based on \textit{Cosine} distance, which makes the clustering results unpredictable when dealing with data in the general $\textit{l}_2$-space. 

In this paper, a new objective function is derived directly from Eqn.~\ref{eqn:tkm}, which makes it suitable for the whole $\textit{l}_2$-space. In other word, objective function proposed in~\cite{ml04:zhao} is the special case of our proposed form. Based on the proposed objective function, conventional egg-chicken k-means iteration is revised to a simpler form. On one hand, when applying the revised iteration procedure in direct k-way clustering, k-means is able to reach to considerably lower clustering distortion within only a few rounds. On the other hand, as the iteration procedure is undertaken in top-down hierarchical clustering manner (specifically bisecting), it shows faster speed while maintaining relatively lower clustering distortion in comparison to traditional k-means and most of its variants.
\section{Clustering Objective Functions}
\label{sec:func}
In this section, the clustering objective functions upon which our k-means method is built are presented. Basically, two objective functions that aim to optimize the clustering results from different aspects are derived. Furthermore, we also show that these two objective functions can be reduced to a single form.

\subsection{Preliminaries}
\label{sec:basic}
In order to facilitate the discussions that are followed, several variables are defined. Throughout the paper, the size of input data is given as \textit{n}, while the number of clusters to be produced is given as \textit{k}. The partition formed by a clustering method is represented as $\{S_1,\cdots,S_r\cdots,S_k\}$. Accordingly, the sizes of clusters are given as $n_1,\cdots,n_r,\cdots,n_k$. The composite vector of a cluster is defined as $D_r=\sum_{x_i \in S_r}x_i$. The cluster centroid $C_r$\footnote{We refer to as column vector across the paper.} is defined by its members,
\begin{equation}
C_r=\frac{\sum_{i=1}^{n_r}{x_i}}{n_r}=\frac{D_r}{n_r}
\end{equation}
The inner-product of $C_r$ is given by $C_r'C_r=\frac{(\sum_{i=1}^{n_r}{x_i})'(\sum_{i=1}^{n_r}{x_i})}{n_r^2}$, which is expanded as following form.
\begin{equation}
     \begin{aligned}[left]
      C_r'C_r&=\frac{1}{n_r^2}[(x_1'x_1+\cdots+x_1'x_i+\cdots+x_1'x_{n_r})+ \\
       &(x_2'x_1+\cdots+x_2'x_i+\cdots+x_2'x_{n_r})+ \\
       &\cdots \\
       &(x_i'x_1+\cdots+x_i'x_i+\cdots+x_i'x_{n_r})+\\
       &\cdots \\
       &(x_{n_r}'x_1+\cdots+x_n'x_i+\cdots+x_{n_r}'x_{n_r})] \\
       \\
           &=\frac{1}{n_r^2}(\sum_{i=1}^{n_r}{x_i^2}+2\sum_{i,j=1 \& i<j}^{n_r}<x_i, x_j>) \nonumber
     \end{aligned}
\end{equation}

Re-arrange the above equation, we have
\begin{equation}
\sum_{i,j=1 \& i<j}^{n_r}<x_i, x_j>=\frac{1}{2}({n_r}^2{\cdot}C_r'C_r-\sum_{i=1}^{n_r}{x_i^2}).
\label{eqn:cross}
\end{equation}
The sum of pairwise $\textit{l}_2$-distance within one cluster is given as
\begin{equation}
 S=(n_r-1)\sum_{i=1}^{n_r}{x_i^2}-2{\cdot}\sum_{i,j=1 \& i<j}^{n_r}<x_i, x_j>.
\label{eqn:sum}
\end{equation}

Plug Eqn.~\ref{eqn:cross} into Eqn.~\ref{eqn:sum}, we have
\begin{equation}
    \begin{aligned}
    S &=({n_r}-1)\sum_{i=1}^{n_r}{x_i^2}-({n_r}^2{\cdot}C_r'C_r-\sum_{i=1}^{n_r}{x_i^2}) \\
        &=({n_r}-1)\sum_{i=1}^{n_r}{x_i^2}-{n_r}^2{\cdot}C_r'C_r+\sum_{i=1}^{n_r}{x_i^2} \\
        &={n_r}\sum_{i=1}^{n_r}{x_i^2}-{n_r}^2{\cdot}C_r'C_r.
    \end{aligned}
    \label{eqn:ssum}
\end{equation}
Eqn.~\ref{eqn:ssum} is rewritten as
\begin{equation}
    S ={n_r}\sum_{i=1}^{n_r}{x_i^2}-D_r'D_r.
    \label{eqn:ssum2}
\end{equation}

\subsection{Objective Functions}
In this section, two objective functions (also known as criterion functions~\cite{ml04:zhao}) are developed. In addition, with the support of the results obtained in Section~\ref{sec:basic}, these objective functions will be reduced to simple forms, which enable them to be carried out efficiently in the incremental optimization procedure.

According to~\cite{ml04:zhao}, objective functions are categorized into two groups. One group of the functions considers the tightness of clusters, while another focuses on alienating different clusters. In this paper, we focus on producing a clustering solution defined over the elements within each cluster. It therefore does not consider the relationship between the elements assigned to different clusters. 

The first objective function we consider is to minimize the distance of each element to its cluster centroid, which is nothing more than the objective function of k-means.
\begin{equation}
 \begin{aligned}
 \mbox{Min. } \mathcal{I}_1 = &\sum_{q(x_i)=r}{\parallel C_r -x_i \parallel^2} \\
                            = &\sum_{r=1}^{k}\sum_{x_i \in S_r}d(x_i,C_r).
  \end{aligned}
 \label{eqn:iz1}
\end{equation}
The above equation is simplified as 
\begin{equation}
 \begin{aligned}
 \mbox{Min. } \mathcal{I}_1 = &\sum_{r=1}^{k}(\sum_{i=1}^{n_r}x_i'x_i+{n_r}{C_r'C_r}-2\sum_{i=1}^{n_r}x_i'C_r) \\
                                = & \sum_{r=1}^{k}(\sum_{i=1}^{n_r}x_i'x_i+\frac{D_r'D_r}{{n_r}}-2\frac{D_r'D_r}{n_r}) \\
                                = & \sum_{r=1}^{k}(\sum_{i=1}^{n_r}x_i'x_i-\frac{D_r'D_r}{n_r}) \\
                                = & \sum_{r=1}^{k}\sum_{i=1}^{n_r}x_i'x_i-\sum_{r=1}^{k}\frac{D_r'D_r}{n_r} \\
                                = & E-\sum_{r=1}^{k}\frac{D_r'D_r}{n_r}
  \end{aligned}
 \label{eqn:iz2}
\end{equation}
Since the input data are fixed, $E$ is a constant. As a result, minimizing Eqn.~\ref{eqn:iz2} is equivalent to maximizing following function
\begin{equation}
        \mbox{Max. } \mathcal{I}^*_1 = \sum_{r=1}^k\frac{D_r'D_r}{n_r}.
        \label{eqn:iz3}
\end{equation}
Although objective function in Eqn.~\ref{eqn:iz3} is in the same form as the first objective function in~\cite{ml04:zhao}, they are derived from different initial objectives. More importantly, in our case, there is no constraint that input data should be in unit length.

The second internal objective function that we will study minimizes the sum of the average pairwise distance between the elements assigned to each cluster, weighted according to the size of each cluster.
\begin{equation}
        \mbox{Min. } \mathcal{I}_2=\sum_{r=1}^k{n_r}(\frac{2}{n_r{\cdot}(n_r-1)}\sum_{d_i,d_j \in S_r \& i > j}d(x_i, x_j))
        \label{eqn:ii1}
\end{equation}

Plug Eqn.~\ref{eqn:ssum2} in, we have
\begin{equation}
        \begin{aligned}
        \mbox{Min. } \mathcal{I}_2&=\sum_{r=1}^k{n_r}(\frac{2}{n_r{\cdot}(n_r-1)}({n_r}\sum_{i=1}^{n_r}{x_i'x_i}-D_r'D_r)) \\
        &=\sum_{r=1}^k\frac{2n_r}{n_r-1}\sum_{i=1}^{n_r}{x_i'x_i}-2\sum_{r=1}^k\frac{D_r'D_r}{n_r-1}
\end{aligned}
        \label{eqn:ii2}
\end{equation}

In Eqn.~\ref{eqn:ii2}, $\frac{n_r}{n_r-1}$ is close to \textit{1}, the above criterion function can be approximated as
\begin{equation}
        \mbox{Min. } \mathcal{I}_2 \approx 2E - 2\sum_{r=1}^k\frac{D_r'D_r}{n_r}.
\label{eqn:ii3}
\end{equation}
Similar as Eqn.~\ref{eqn:iz2}, since the input data are fixed, $E$ is a constant. As as result, minimizing Eqn.~\ref{eqn:ii3} is equivalent to maximizing function
\begin{equation}
        \mbox{Max. } \mathcal{I}^*_2 \approx \sum_{r=1}^k\frac{D_r'D_r}{n_r}.
        \label{eqn:ii4}
\end{equation}

Noticed that similar optimization objectives have been discussed under \textit{Cosine} similarity measure in~\cite{ml04:zhao}. As it is shown that above two objective functions are the same in the paper. This is different from the result obtained in our case (general $l_2$-space). As shown above, in $l_2$-space, the objective functions for $\mathcal{I}^*_1$ and $\mathcal{I}^*_2$ are only approximately the same. The advantage that two objective functions are reduced to the same form is that, when we try to optimize one objective function, we optimize another in the mean time. Specifically, when we minimize the distances from elements to their cluster centroid, the average intra-cluster distance is minimized in the meantime. Since these two objective functions can be simplified to the same form, only objective function $\mathcal{I}^*_1$ is discussed in the rest of paper.

Although objective function in Eqn.~\ref{eqn:iz3} is derived from Eqn.~\ref{eqn:tkm}, the former is much easier to operate in the incremental k-means procedure. As it will be shown in the next section, it is quite convenient to evaluate whether Eqn.~\ref{eqn:iz3} attains a higher score (implies lower distortion in terms of Eqn.~\ref{eqn:tkm}) when a sample $x_i$ is moved from one cluster to another.
\section{K-means Driven by Objective Function}
\label{sec:alg}
In this section, with the objective function developed in Section~\ref{sec:func}, two iterative clustering procedures are presented. Namely, one produces \textit{k} clusters directly (called as direct \textit{k}-way k-means), while another produces \textit{k} clusters by bisecting input data sequentially \textit{k-1} times (called as bisecting k-means). Both  clustering strategies are built upon incremental clustering~\cite{ml04:zhao,pttnc01} and driven by objective function $\mathcal{I}^*_1$ (Eqn.~\ref{eqn:iz3}). 

\subsection{Clustering Algorithm}
The basic idea of incremental clustering is that one sample $x_i$ is moved from cluster $S_u$ to $S_v$ as soon as this movement leads to higher score of objective function $\mathcal{I}^*_1$. To facilitate our discussion, the new function value as sample $x_i$ is moved from $S_u$ to $S_v$ is formulated as following.
\begin{equation}
\begin{aligned}
   \mathcal{I}^*_1(x_i)=&\frac{(D_v+x_i)'(D_v+x_i)}{n_v+1}+\frac{(D_u-x_i)'(D_u-x_i)}{n_u-1} \\
   =&\frac{D_v'D_v+2x_i'D_v+x_i'x_i}{n_v+1}+\frac{D_u'D_u-2x_i'D_u+x_i'x_i}{n_u-1}\\
   =&2x_i'\frac{D_v}{n_v+1}-2x_i'\frac{D_u}{n_u-1}+\frac{D_v'D_v}{n_v+1}+\frac{D_u'D_u}{n_u-1}+\frac{x_i'x_i}{n_v+1}+\frac{x_i'x_i}{n_u-1}
\end{aligned}
\label{eqn:val}
\end{equation}
In each iteration of the clustering, sample $x_i$ is randomly selected. The algorithm checks whether moving $x_i$ from its current cluster to any other cluster will lead to higher $\mathcal{I}^*_1$ (i.e., ${\Delta}\mathcal{I}^*_1 > 0$). If it is the case, $x_i$ is moved to another cluster. The clustering procedure is detailed in Alg.~\ref{alg:kway}.
\begin{figure*}
\begin{center}
        \subfigure[initialization]
        {\includegraphics[width=0.26\linewidth]{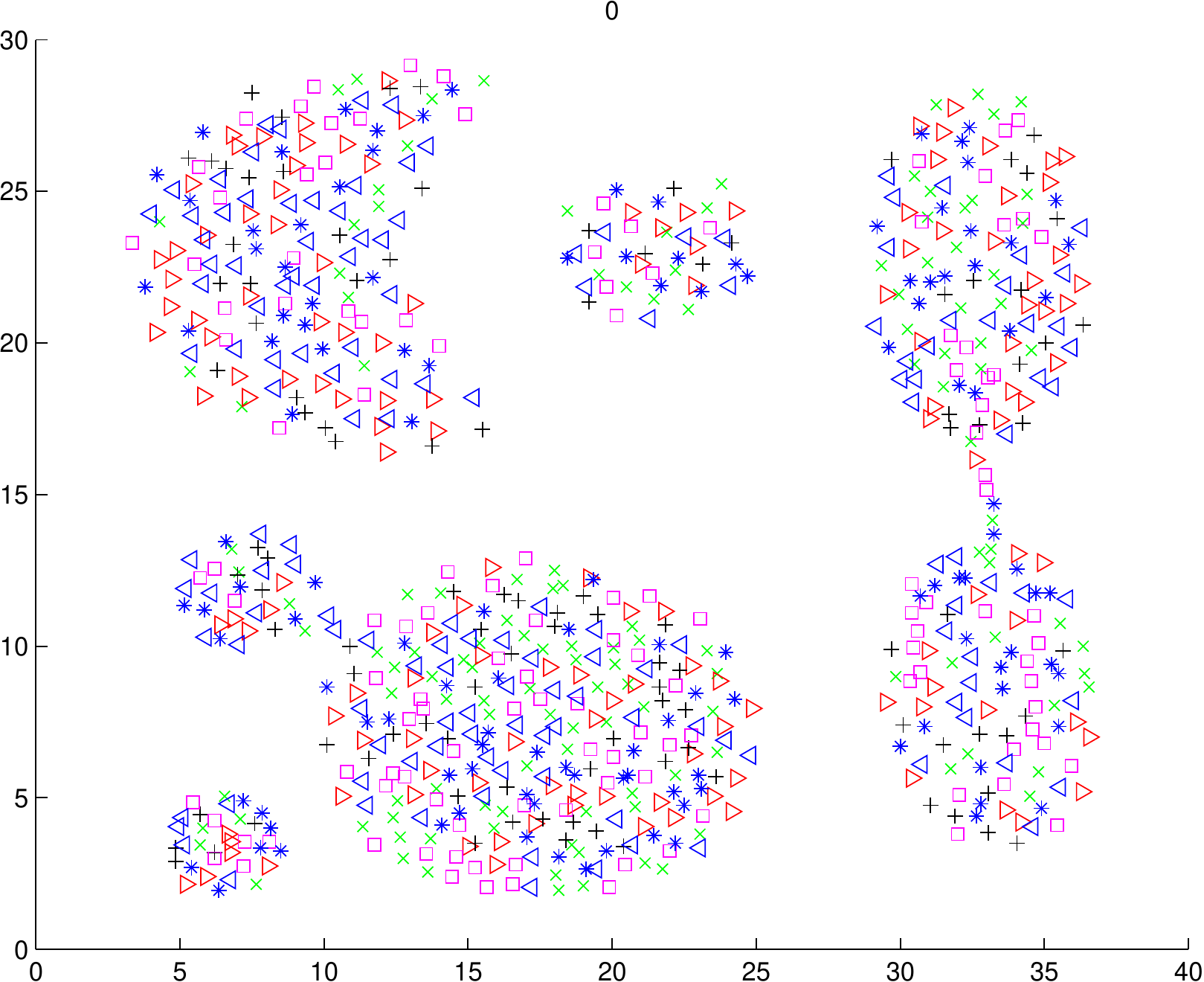}}
        \hspace{0.14in}
        \subfigure[iter=1]
        {\includegraphics[width=0.26\linewidth]{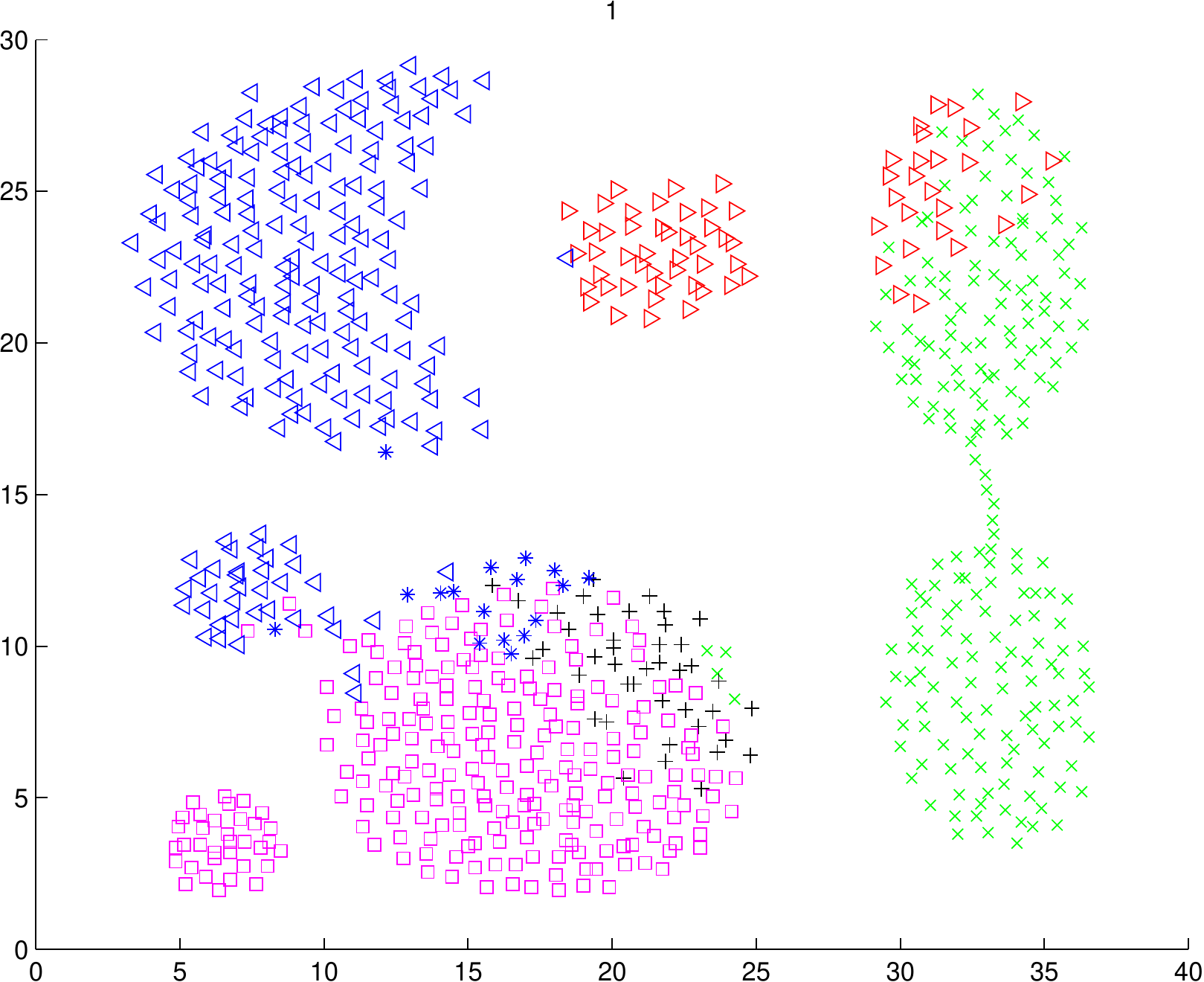}}
        \hspace{0.14in}
        \subfigure[iter=10]
        {\includegraphics[width=0.26\linewidth]{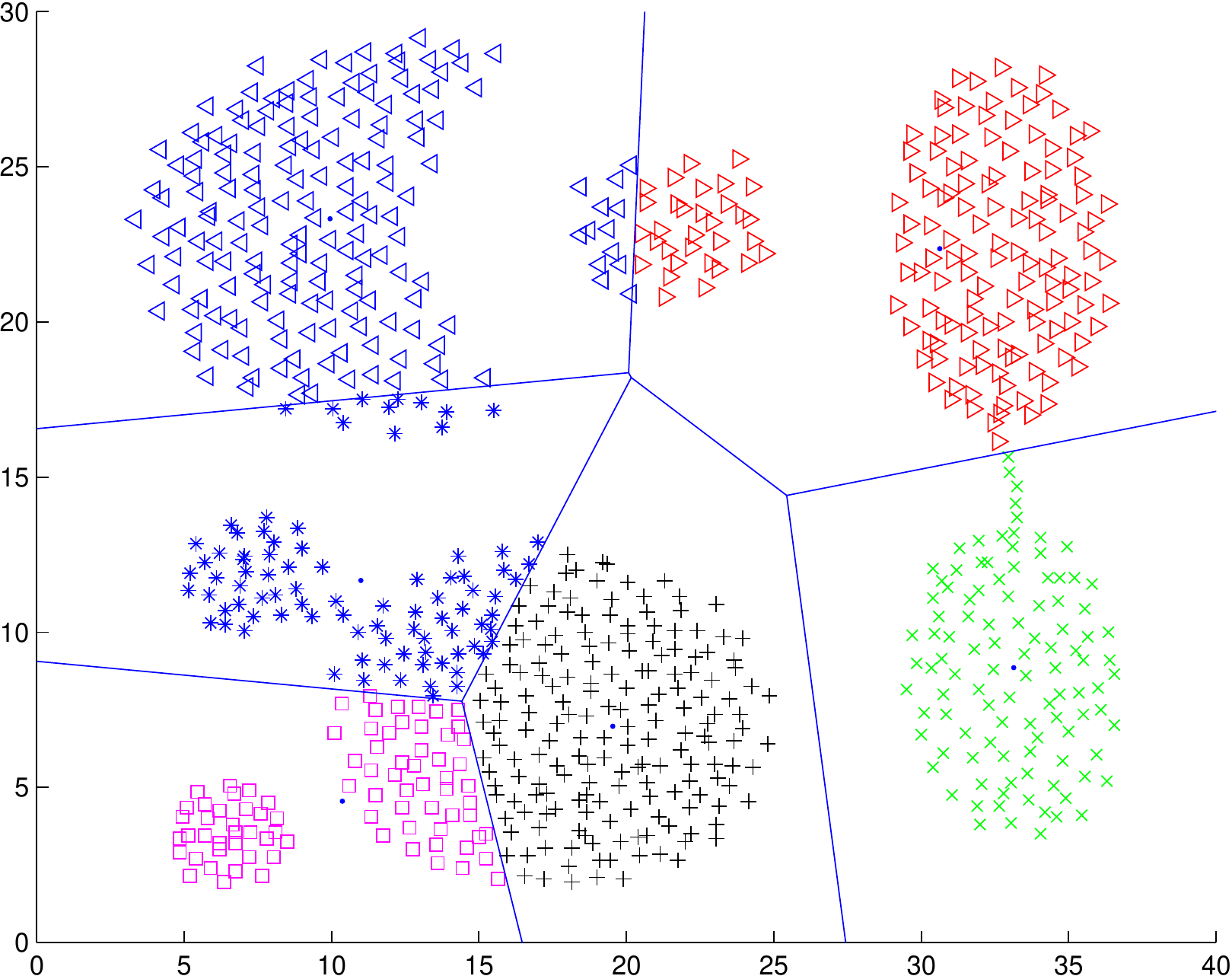}} 
\caption{Illustration of direct k-way k-means clustering with Alg.~\ref{alg:kway}. The clustering process starts from the state that samples are all assigned with random label. The final cluster centroids in (c) form a convex partition over the 2D space, which are called as \textit{Voronoi} diagram. According to \textit{Lloyd}'s condition, all the samples belonging to one cluster fall into the same \textit{Voronoi} cell.}
\label{fig:xtkdemo}
\end{center}
\end{figure*}

As seen from \textbf{Step 3} of Alg.~\ref{alg:kway}, the initialization of our method is different from most of the current practice of k-means, there is no assignment of each sample to its closest initial centroid. On the contrary, each sample $x_i$ is assigned with a random cluster label (ranges from \textit{1} to \textit{k}). This allows to calculate an initial score of $\mathcal{I}^*_1$ and the composite vector \textit{D} of each cluster. It is possible to do the initial assignment following the way of k-means or k-means++~\cite{kpp07}. However, as will be revealed in Section~\ref{sec:exp}, initialization under either k-means manner or k-means++ manner improves the clustering quality slightly. However, extra computation is required in such kind of initial assignment. 

During each iteration, each sample $x_i \in X$ is checked in random order. The optimization in \textbf{Step 8}-\textbf{10} seeks the movement of $x_i$ that leads to highest increase of function score. From the optimization point of view, the algorithm reduces the clustering distortion greedily. From another point of view, the seeking process is comparable to the sample-to-centroid assignment in traditional k-means. They are actually on the same computational complexity level. 

Whereas it is not necessary that we must seek the best movement for $x_i$. As we discover by experiment, it is feasible that moving $x_i$ to another cluster as long as we find ${\Delta}\mathcal{I}^*_1(x_i)$ is greater than \textit{0}. On one hand, this will speed-up the iteration. On the other hand such kind of scheme usually takes more rounds to reach to the same level of distortion. However, we discover that such kind of less greedy scheme results in lower clustering distortion if the iteration loops for sufficient number of times.

Moving $x_i$ from one cluster to another (\textbf{Step 9}) is very convenient to take. It includes the operation that updates the cluster label of $x_i$ and the operation that updates the composite vector for cluster $S_v$ and $S_u$, viz., $D_v=D_v+x_i$, $D_u=D_u-x_i$.

Note that this incremental updating scheme is essentially different from online learning vector quantization (LVQ)~\cite{map01:kohonen}, in which the cluster centroids are updated incrementally. In the above iteration procedure, no cluster centroids are explicitly produced. As a result, there is no need to update cluster centroid. The clustering iteration is explicitly driven by an objective function rather than driven by the discrepancy between cluster centroids and their cluster members. As revealed later in the experiment, compared to LVQ, Alg~\ref{alg:kway} is more efficient and leads to considerably lower distortion.
\begin{algorithm}{Direct \textit{k}-way Boost k-means}
  \begin{algorithmic}[1]
    \STATE \textbf{Input}: matrix $X_{n{\times}d}$
    \STATE \textbf{Output}: $S_1,{\cdots},S_r,{\cdots}S_k$
    \STATE Assign $x_i \in X$ with a random cluster label;
    \STATE Calculate $D_1,{\cdots},D_r,{\cdots}D_k$ and $\mathcal{I}^*_1$;
    \WHILE {not convergence}
    \FOR {each $x_i \in X$ (in random order)}
    \STATE Seek $S_v$ that maximizes ${\Delta}\mathcal{I}^*_1(x_i)$;
    \IF {${\Delta}\mathcal{I}^*_1(x_i) > 0$}
    \STATE Move $x_i$ from current cluster to $S_v$;
    \ENDIF
    \ENDFOR
    \ENDWHILE
  \end{algorithmic}
  \label{alg:kway}
\end{algorithm}
Figure~\ref{fig:xtkdemo} illustrates three iterations of Alg.~\ref{alg:kway} in 2D case. As shown in the figure, the initial clutsering result is random and messy. Samples belonging to different clusters are totally mixed up. However, only after one round of iteration, the clustering result becomes much more compact. The clustering terminates at the \textit{10}th round, where \textit{Lloyd}'s condition is reached. The optimality of this procedure is analyzed in Appendix~\textbf{A} and its convergence is proved in Appendix~\textbf{B}.

Overall, method presented in Alg.~\ref{alg:kway} is different from traditional k-means in three major aspects. Firstly, no initial assignment is required. Moreover, the egg-chicken loop in the traditional k-means has been replaced by a simpler stochastic optimization procedure. Furthermore, unlike traditional k-means, it is not necessary to seek the best movement for each sample in the iteration. 

The method presented in Alg.~\ref{alg:kway} is on the same complexity level as traditional k-means (i.e., $O(t{\cdot}n{\cdot}d{\cdot}k)$), which is unbearably slow when dealing with large-scale data. 

The method is revised into a top-down hirarchical clustering version for large-scale clustering. Specifically, at each time, one intermediate cluster is selected and bisected into two smaller clusters by calling Alg.~\ref{alg:kway}. The details of this method are given in Alg.~\ref{alg:bsect}.

As shown in Alg.~\ref{alg:bsect}, priority queue \textit{Q} pops out one cluster for bisecting each time. As discussed in~\cite{kddzhao05}, there are basically two ways to organize the priority queue. One can prioritze the cluster with biggest size or the one with highest average intra-cluster distance to split. Similar as~\cite{kddzhao05}, we find splitting the biggest cluster usually demonstrates more stable performance. As a result, the queue is sorted in descending order by the cluster sizes in our practice.

\begin{algorithm}{Bisecting Boost k-means}
  \begin{algorithmic}[1]
    \STATE \textbf{Input}: matrix $X_{n{\times}d}$
    \STATE \textbf{Output}: $S_1,{\cdots},S_r,{\cdots}S_k$
    \STATE Treat \textit{X} as one cluster $S_1$;
    \STATE Push $S_1$ into a priority queue \textit{Q};
    \STATE i = 1;
    \WHILE {$i < k$}
    \STATE Pop cluster $S_i$ from queue \textit{Q}
    \STATE Call Alg.~\ref{alg:kway} to bisect $S_i$ into $\{S_{i}, S_{i+1}\}$;
    \STATE Push $S_{i}, S_{i+1}$ into queue \textit{Q};
    \STATE i = i + 1;
    \ENDWHILE
  \end{algorithmic}
  \label{alg:bsect}
\end{algorithm}

It is possible to partition the intermediate cluster into more than two clusters each time. In the following, we are going to show that this bisecting scheme achieves highest scalability among all alternative top-down secting schemes.
\subsection{Scalability Analysis}
\label{sec:alg2}
In this section, the computation complexity of Alg.~\ref{alg:bsect} is studied by considering the total number of comparisons required in the series of bisecting clustering. The number of iterations in each bisecting is assumed to be a constant by taking the average number of iterations.

In order to facilitate the analysis while without loss of generality, we assume that each intermediate cluster in Alg.~\ref{alg:bsect} is partitioned evenly. In addition, we generalize Alg.~\ref{alg:bsect} to an s-secting algorithm. Namely, an intermediate cluster is partitioned to \textit{s} ($s \ge 2$) clusters. Now we consider the size of series of intermediate clusters that are produced when performing sequential secting. Given \textit{q} is the depth of splitting, it is easy to see ${\lceil}\log_s{k}{\rceil}=q+1$. The sizes of all intermediate clusters are given as following.
\begin{equation}
\begin{aligned}
n,\underbrace{ \frac{n}{s}, \frac{n}{s},..}_s, \underbrace{\frac{n}{s^2}, \frac{n}{s^2},...}_{s^2},  ..... , \underbrace{\frac{n}{s^q}, \frac{n}{s^q},...}_{s^q} \nonumber\\
\end{aligned}
\end{equation}
As a result, the number of samples to be visited during the clustering procedure is
\begin{equation}
\begin{aligned}
&n+\frac{n}{s}*s^1+\frac{n}{s^2}*s^2+\frac{n}{s^3}*s^3.....+\frac{n}{s^q}*s^{q}\\
= &n+\underbrace{n + n + n + ... + n}_{q}\\
= &n*(1+q)\\
\approx &n*\log_{s}{k}.
\end{aligned}
\end{equation}
Considering that one sample has to compare with $s-1$ centroids each time, the total number of comparisons is
\begin{equation}
\begin{aligned}
n*(s-1)*\log_{s}k.
\end{aligned}
\label{eqn:comptm}
\end{equation}
Given \textit{n} and \textit{k} are fixed, Eqn.~\ref{eqn:comptm} increases monotonically with respect to \textit{s}. As a result, the number of comparisons reaches to the minimum (i.e., $n\log_{2}k$) when $s=2$. To this end, it is clear that bisecting is the most efficient secting scheme.

Compared with Alg.~\ref{alg:kway}, the complexity of Alg.~\ref{alg:bsect} is reduced to $O(\bar{t}{\cdot}n{\cdot}d{\cdot}log(k))$, where $\bar{t}$ is the average number of iterations in each bisecting. Compared with \textit{t} in traditional k-means, $\bar{t}$ is much smaller given the scale of clustering problem is much smaller in terms of both the size of input data and the number of clusters to be produced. As a result, the complexity of  Alg.~\ref{alg:kway} has been largely reduced since term $n{\cdot}d$ has been multiplied by a much smaller factor $\bar{t}{\cdot}log(k)$.
\begin{figure}[t]
\begin{center}
        \includegraphics[width=0.850\linewidth]{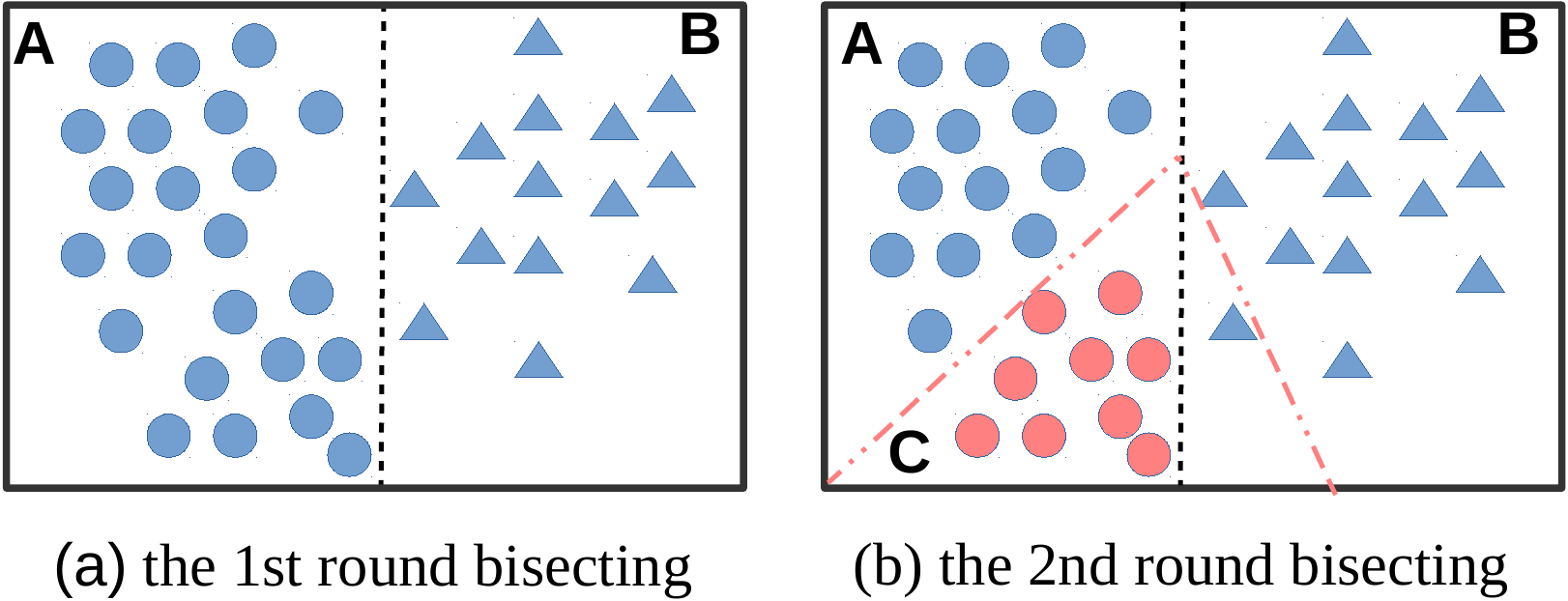}
\caption{Illutration of two consecutive bisecting in the bisecting clustering where \textit{Lloyd}'s condition breaks.}
\label{fig:lloyddemo}
\end{center}
\vspace{-0.25in}
\end{figure}

Although Alg.~\ref{alg:bsect} is efficient, the clustering result produced by Alg.~\ref{alg:bsect} unfortunately does not satisfy with \textit{Lloyd}'s condition. This problem is illustrated in Figure~\ref{fig:lloyddemo}. As one of the clusters is further partitioned into two (from Figure~\ref{fig:lloyddemo}(a) to Figure~\ref{fig:lloyddemo}(b)), the partition over 2D space is formed by centroids changes. Cluster C claims bordering points from cluster B. However, points from cluster B cannot be reassigned to cluster C if no further intervention is involved. This is actually an underfitting issue and exists for any hierarchical clustering method. Fortunately, this issue can be alleviated by adopting Alg.~\ref{alg:kway} as a refinement procedure after Alg.~\ref{alg:bsect} outputs \textit{k} clusters. To do so, extra time is required. It therefore becomes a problem of balancing between efficiency and quality.

According to our observation, it is possible to further speed-up the proposed boost k-means. After a few iterations, both k-means and boost k-means will be trapped in a local minima. Only samples that bordering between different clusters should be shuffled from one cluster to another. As a result, given a sample, it is no need to search for the best movement among k clusters. Instead, sample only needs to compare with top-$k_0$ ($k_0 \ll k$) centroids to search the suitable movement. We find that, this simple modification typically leads to $7{\sim}8$ times speed-up while without significant performance degradation.
\section{Experiments}
\label{sec:exp}
\begin{table*}[t]
\begin{center}
\caption{Configurations of k-means and its variants and their corresponding Abbreviations}
\footnotesize{
\begin{tabular}{|c|c|c|c|c|}
\hline
 & \multicolumn{2}{c|}{k-means} & \multicolumn{2}{c|}{boost k-means} \\ \hline
Initial assigment & k-way & bisecting & k-way & bisecting \\ \hline
Random & k-means~\cite{km82} & BsKM & BKM(rnd) & BsBKM(rnd)\\ \hline
Probability based~\cite{kpp07} & k-means++~\cite{kpp07}& BsBKM++  & BKM(kpp) & BsBKM(kpp)\\ \hline
None & - & - & BKM(non) & BsBKM(non) \\ \hline
\end{tabular}
}
\label{tab:conf}
\vspace{-0.2in}
\end{center} 
\end{table*}

In this section, the effectiveness of proposed clustering method, namely boost k-means (BKM) is studied under different scenarios. In the first experiment, dataset SIFT1M~\cite{JDS11} is adopted to evaluate the clustering quality. In the second experiment, BKM is tested on the nearest neighbor search task based on product quantizer (PQ)~\cite{JDS11} in which this method is adopted for quantizer training. In the third experiment, BKM has been applied to traditional document clustering. Following the practice of~\cite{ml04:zhao,kddzhao05}, \textit{15} document datasets\footnote{Available at http://glaros.dtc.umn.edu/gkhome/fetch/sw/cluto/datasets.tar.gz} have been adopted. In the last experiment, the scalability of BKM has been tested on large-scale image clustering task, for which the number of images we use is as large as \textit{10} million.

In our study, the performance from traditional k-means is treated as comparison baseline. In addition, representative k-means variants, such as Mini-Batch~\cite{mnkm10}, Repeated Bisecting k-means (RBK)~\cite{kddzhao05}, online Learning Vector Quantization (LVQ)~\cite{map01:kohonen} and k-means++~\cite{kpp07} are considered in the comparison. For Mini-Batch, our configuration makes sure that the iteration covers \textit{10\%} of the input data. The configuration is fixed across all the experiments. For RBK, we select the objective function that maximizes the average \textit{Cosine} similarity between samples within one cluster, which is the special case of ours given the input data is $\textit{l}_2$-normalized. LVQ is similar to k-means except that in each round, a cluster centroid is upated as soon as a sample is assigned. The updating rate starts from \textit{0.01} and decreases at a pace of $4\times10^{-4}$ in one iteration.

As shown in Table~\ref{tab:conf}, there are variants of k-means depending on cluster initialization and data partitioning strategies (e.g., direct k-way or bisecting). This is also true for the proposed BKM. In the table, `initial assignment' refers to the operation of assigning each sample to its closest initial centroid. When the initial assignment is based on random seeding like traditional k-means, it is denoted as `rnd'. When it is based on probability distribution seeding as k-means++, it is denoted as `kpp'. Initialization without initial assignment is denoted as `non'. In the experiments, all the variants out of these different configurations on k-means as well as BKM are considered. Performance evaluation is separately conducted for k-way and bisecting clustering method. Noted that BsBKM(rnd) is the same as RBK if the input data is $\textit{l}_2$-normalized. The experiment in this section is conducted using 1 million SIFT features~\cite{Low04}. The features are clustered into \textit{10,000} partitions and the average distortion error is calculated for performance evaluation.

In addition, we also study the performance trend of BKM when \textbf{Steps} \textbf{7}-\textbf{10} in Alg.~\ref{alg:kway} are modified to moving the sample as soon as ${\Delta}I_1(x_i)>0$. The variants under this modification are denoted as BKM(xxx)+Fast.~\footnote{Note that this is not applicable for bisecting BKM.} All the methods considered in the paper are implemented in C++ and the simulations are conducted on a PC with \textit{2.4}GHz Xeon CPU and \textit{32}G memory setup.

\subsection{Evaluation of Clustering Distortion}
\label{sec:distort}
Since k-means and most of its variants share the same objective function (Eqn.~\ref{eqn:tkm}), it is straightforward to evaluate the clustering performance by checking to what degree the objective is reached. The average distortion (given in Eqn.~\ref{eqn:distor}) is adopted for evaluation~\cite{ikmn15}, which takes average over Eqn.~\ref{eqn:tkm},
\begin{equation}
         \footnotesize{
\mathcal{E} = \frac{\sum_{q(x_i)=r}{\parallel C_r -x_i \parallel^2}}{n}.
}
\label{eqn:distor}
\end{equation}
For above equation, the lower the distortion value, the better quality of the clustering result is.

\begin{figure}
\begin{center}
	\subfigure[variations in initialization]
    {\includegraphics[width=0.4\linewidth]{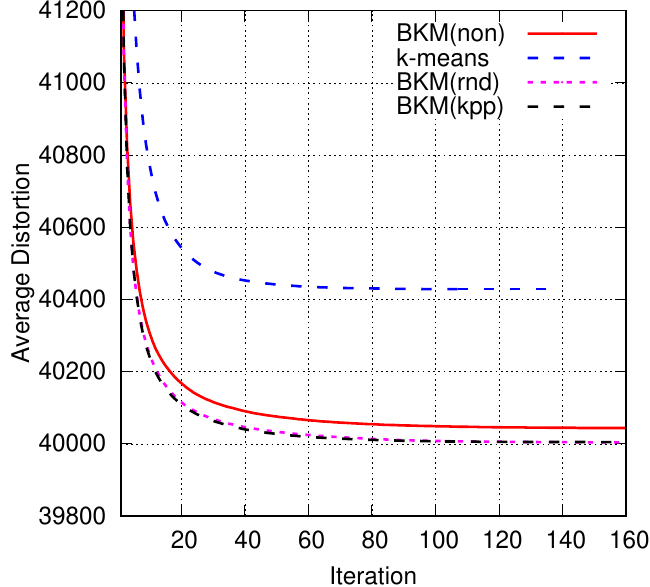}}
    \hspace{0.025in}
	\subfigure[adhoc versus best movements]
    {\includegraphics[width=0.4\linewidth]{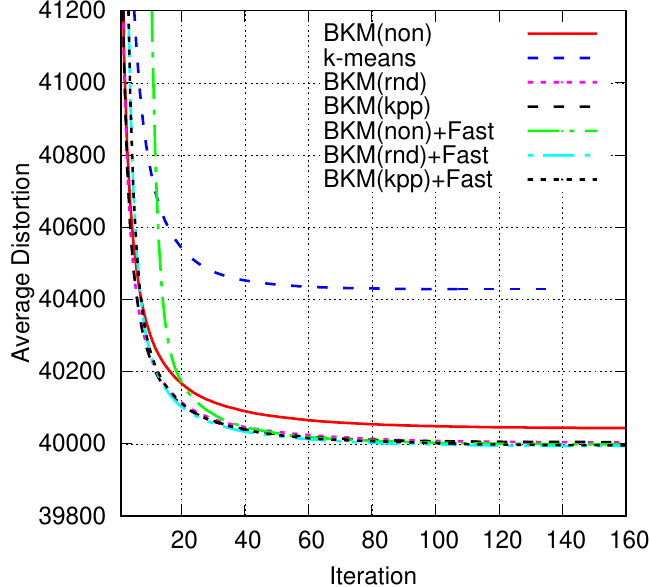}} \\
    	\subfigure[comparison with k-means variants]
    {\includegraphics[width=0.4\linewidth]{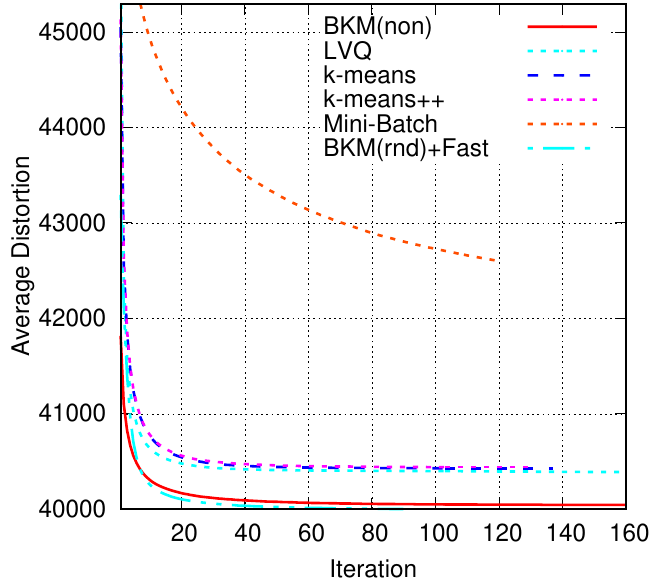}}
    \hspace{0.025in}
    \subfigure[significance test on SIFT100K]
    {\includegraphics[width=0.4\linewidth]{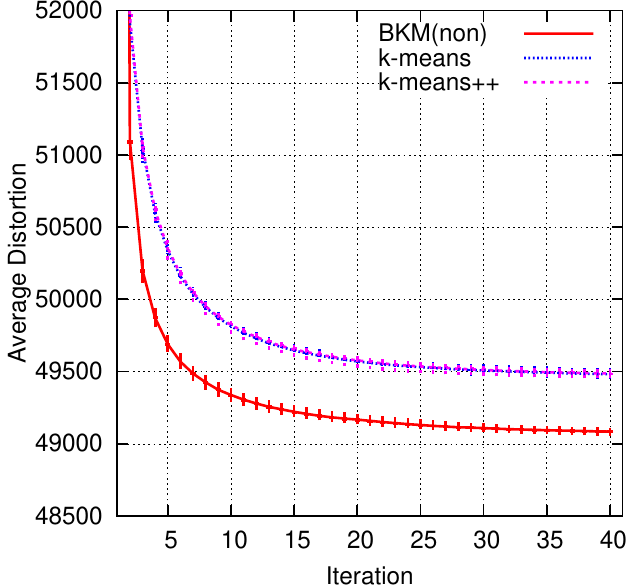}}
    \caption{The experiments are conducted on SIFT1M for figures (a)-(c) and on SIFT100K for figure (d). The results show different performance: (a) impact of initialization in different ways; (b) fast version of BKM by not seeking optimal movement in the steps 7-10 of Alg~1; (c) Comparison of BKM to variants of k-means; (d) significance of improvement over other k-means variants achieved by BKM by repeating the experiments by 128 runs.}
\label{fig:distort}
\end{center}
\end{figure}

The first experiment mainly studies the behavior of the proposed BKM under different initializations. The average distortion curves produced by variants direct k-way BKM are given in Figure~\ref{fig:distort}(a) as a function of numbers of iteration. Traditional k-means is treated as baseline for performance comparison. The result shows that clustering distortion of BKM drops faster than traditional k-means. The average distortion from traditional k-means is around \textit{40,450} after \textit{130} iterations. In contrast, BKM without initial assignment (BKM(non)) is able to reach to the same distortion level after only \textit{7} iterations. Moreover, we find that initializing BKM as traditional k-means way (BKM(rnd)) or as k-means++ (BKM(kpp)) allows the iteration to start from a low distortion level. Nevertheless the advantage over BKM(non) fades away after \textit{15} iterations. In comparison to BKM(non), the extra cost of adopting initial assignment in BKM is relatively high.

The second experiment studies the performance trend of Alg.~\ref{alg:kway} in case when \textbf{Step}s \textbf{7}-\textbf{10} do not seek the best movement (BKM(xxx)+Fast). As shown in Figure~\ref{fig:distort}(b), the distortion drops slower than BKM(non) which seeks the best movement. However, lower distortion is achievable by BKM(rnd)+Fast when reaching to sufficiently number of iterations (e.g., \textit{20} iterations). This indicates that when the optimization scheme is more greedy, it is likely to get trapped in a worse local optima. This observation applies to BKM under different kinds of initialization. Noted that the time cost for BKM(xxx)+Fast is lower than that of BKM that seeks the best movement in each iteration. Whereas, BKM(xxx)+Fast usually needs a few more number of iterations to reach to the similar distortion level. Overall, as investigated in Section~\ref{sec:tm}, BKM(xxx)+Fast is $\%5$  faster than BKM(xxx). 

Figure~\ref{fig:distort}(c) studies the trend of average distortion among the proposed BKM (specifically BKM(non)), traditional k-means, k-means++, Mini-Batch and LVQ. For all the methods presented, their distortion decreases steadily as the iteration continues. A big performance gap is observed between Mini-Batch and other k-means variants. In addition k-means and k-means++ share similar distortion curve. BKM(non) outperforms k-means and k-means++ by requiring only \textit{7} iterations. Most of the methods including k-means and k-means++ take more than \textit{120} iterations to finally converge. On the other hand, little distortion is observed after \textit{20} iterations, which implies the possibility of terminating the iteration at \textit{20}.  Although similiar as BKM, LVQ updates the intermediate clusters incrementally, updating cluster centroid directly turns out to be inefficient, which leads to considerably poor performance.

Since k-means and its variants are all sensitive to initialization, the performance fluctuates from one run to another. The candelstick chart shown in Figure~\ref{fig:distort}(d) further confirms the significance of the improvement achieved by BKM. This chart is plotted with \textit{128} clustering runs ($k=1,024$) on SIFT100K~\cite{JDS11} for each method. As shown in the figure, although the performance flucturates for all the methods, the variations are minor. Similar as previous observation, there is no significant difference between traditional k-means and k-means++. In contrast,  the performance gap between BKM and traditional k-means is much more significant than the performance variations across different runs.

Table~\ref{tab:distortbi} shows the average distortion of different k-means variants under bisecting strategy. The result from k-means (after \textit{130} iterations) is presented for the comparison. As shown from the table, the average distortion from all bisecting methods are on the level of \textit{$4.5{\times}10^4$}. Methods built upon~Alg.~\ref{alg:kway} always perform better. The average distortion from all bisecting clustering methods are much higher than that of k-means. They are actually only close to the distortion level of k-means after one iteration. However, the merit of clustering with bisecting strategy is that it is more than \textit{20} times faster than k-means of a single iteration. The relatively poor clustering quality produced by bisecting strategy is mainly due to the issue of underfitting (as discussed in Section~\ref{sec:alg2}). The clustering results can be further refined by Alg.~\ref{alg:kway} as shown on the \textit{3rd} row of Table~\ref{tab:distortbi}. 

\begin{table*}[t]
\begin{center}
\caption{Average Distortion from K-means Variants under Bisecting Strategy}
\footnotesize{
\begin{tabular}{|c|c||c|c|c|c|c|c|}
\hline
Method & k-means  & RBK & BsKM & BsKM++ & BsBKM(non) & BsBKM(rnd) & BsBKM(kpp) \\ \hline
$\mathcal{E}$ & 40,450.0 & 45,713.5 & 45,835.2 & 45,823.8 & \textbf{45,650.7} & 45,661.2 & 45,658.4 \\ \hline
$\mathcal{E}$ after Rfn. & - & 43,364.4 & 4,3323.9 & 43,366.2 & 43,293.3 & 43,285.5 & \textbf{43,285.4} \\ \hline 
\end{tabular}
}
\label{tab:distortbi}
\end{center}
\end{table*}	

As learned from above experiments, on one hand initial assignment under k-means manner or under k-means++ manner is able to improve the performance of BKM slightly. On the other hand, the initial assignment slows down the method considerably. A trade-off has to be made. In the following experiments, only the results from two representative configurations of BKM, namely BKM(non) and BKM(rnd)+Fast are presented. We leave the other possible configurations to the readers.

\subsection{Nearest Neighbor Search by Product Quantizer (PQ)}
In this section, BKM is applied for visual vocabulary training using product quantization~\cite{JDS11}. Following the practice of~\cite{JDS11}, \textit{100}K SIFT features are used for product quantizer training, while SIFT1M set~\cite{JDS11} is encoded with the trained product quantizers as the reference set for nearest neighbor search (NNS). The obtained recall@top-k is averaged over \textit{1,000} queries for each method. In the experiment, two different settings are tested for product quantizer. Namely,  the \textit{128}-dimensional SIFT vector is encoded with \textit{8} and \textit{16} product quantizers respectively. For clarity, the evaluations are separately conducted for direct k-way and bisecting k-means.

\begin{figure}
\begin{center}
	\subfigure[direct k-way, m=8]
	{\includegraphics[width=0.40\linewidth]{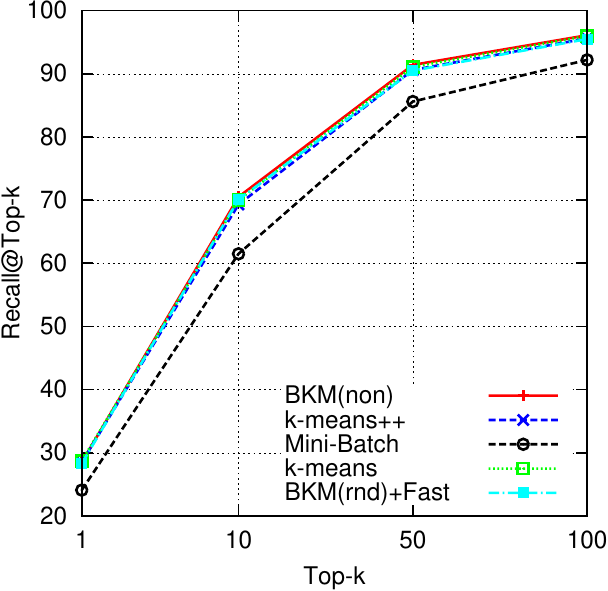}}
	\subfigure[bisecting, m=8]
	{\includegraphics[width=0.385\linewidth]{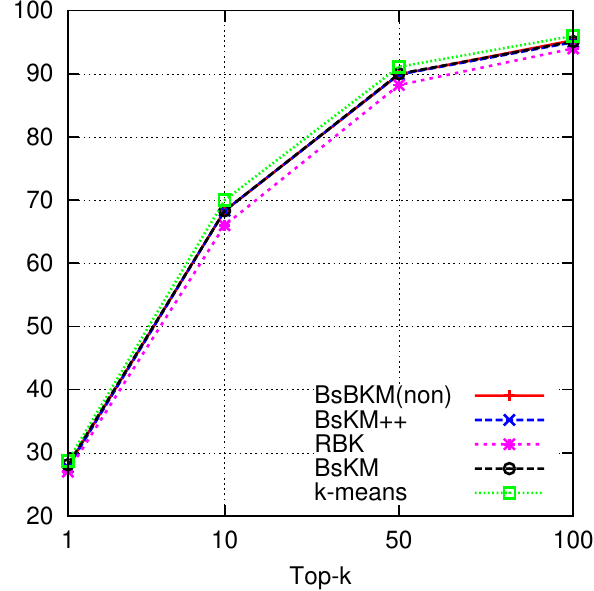}}
	\subfigure[direct k-way, m=16]
	{\includegraphics[width=0.40\linewidth]{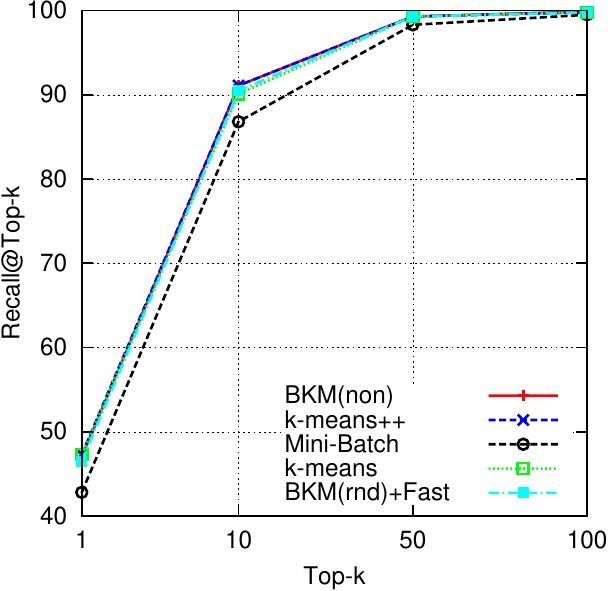}}
	\subfigure[bisecting, m=16]
	{\includegraphics[width=0.385\linewidth]{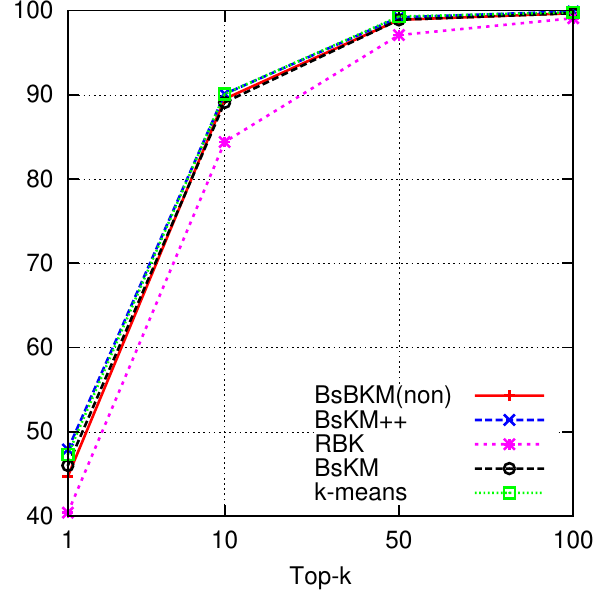}}
\caption{Performance of Nearest Neighbor Search by PQ on SIFT1M when adopting different clustering methods for quantizer training. The size of each product quantizer is fixed to \textit{256} across all the experiments. The assymetric distance calculation (ADC)~\cite{JDS11} is adopted for nearest neighbor search.}
\label{fig:pqtopk}
\vspace{-0.3in}
\end{center}
\end{figure}

Recall@top-100 for direct k-way are presented in Figure~\ref{fig:pqtopk}(a)-(d) under two different settings ($m=8$ and $m=16$).
As seen from the figures, the performances from k-means, k-means++ and BKM(non) are all very close to each other under different settings. The product quantizer trained with bisecting clustering methods shows only \textit{0.1}-\textit{1.3\%} lower performance than that of direct k-way methods.
This basically indicates that product quantizer itself is tolerant to clustering quality. The performance of Mini-Batch and RBK is around \textit{2}-\textit{6\%} lower than the other methods. The poor performance of RBK basically indicates the optimization objective function under \textit{Cosine} similarity is not directly feasible for general $\textit{l}_2$-space.

\subsection{Document Clustering}
\label{sec:expdc}
In this section, the performance of proposed method is evaluated under the context of document clustering. Following in~\cite{ml04:zhao}, \textit{15} document datasets are used for evaluation. The documents has been represented with \textit{TF/IDF} model and normalized to unit length. Similar to~\cite{ml04:zhao}, entropy is adopted for the evaluation as following
\begin{equation}
         \footnotesize{
Entropy = \sum_{r=1}^k\frac{n_r}{n}\frac{1}{\log{c}}*\sum_{i=1}^c{\frac{n_r^i}{n_r}*\log{\frac{n_r^i}{n_r}}},
}
\label{eqn:entropy}
\end{equation}
where \textit{c} is the number of classes. Eqn.~\ref{eqn:entropy} evaluates to what degree that elements from the same class are put in one cluster. The lower the value, the better the result is. In the experiment, each method performs clustering for \textit{10} runs, and the run with the lowest entropy is presented in Table~\ref{tab:entropy}. The presented entropy are averaged over \textit{15} datasets.

\begin{table}
\begin{center}
\caption{Clustering performance (average entropy) on 15 datasets}
\footnotesize{
\begin{tabular}{|c|c|c|c|c|}
\hline
 &  k = 5 &  k = 10  &  k = 15 &  k = 20 \\ \hline
k-means &  0.539 &  0.443 &  0.402 &  0.387 \\ \hline
k-means++ &  0.550 &  0.441 &  0.403 &  0.389 \\ \hline
Mini-Batch &  0.585 &  0.488 &  0.469 &  0.475 \\ \hline
LVQ & 0.800 & 0.761 & 0.681 & 0.674 \\ \hline
BKM(non) &  0.552 &  0.442 &  0.388 &  0.368 \\ \hline
BKM(rnd)+Fast &  \textbf{0.506} &  \textbf{0.419} & \textbf{0.380} &  \textbf{0.353} \\ \hline \hline

BsKM &  0.532 &  0.438 &  0.410 &  0.373 \\ \hline
BsKM++ &  0.507 &  0.422 &  0.400 &  0.379 \\ \hline
BsBKM(non) &  0.514 &  \textbf{0.388} &  \textbf{0.353} & \textbf{0.329} \\ \hline 
RBK & \textbf{0.486} & 0.402 & 0.366 & 0.339 \\ \hline
\end{tabular}
}
\label{tab:entropy}
\end{center}
\end{table}

In general, methods based on BKM perform considerably better. Furthermore, methods with bisecting strategy demonstrate slightly better performance than that of direct k-way in the document clustering task, which shares similar observation as~\cite{kddzhao05}. Overall, BsBKM(non) shows the best performance. While the performance of RBK is close to BsBKM(non). These two methods are quite similar except that no initial assignment is involved in BsBKM(non). This indicates the advantage of no initial assignment in this scenario, which allows clustering to converge to a better local optima.


\subsection{Scalability Test on Image Clustering}
\label{sec:tm}
In this section, the scalability of the proposed k-means is tested on image clustering. The experiment is conducted on \textit{10} million Flickr images (\textit{Flickr10M}), which are a subset of \textit{YFCC100M}~\cite{yfcc}. \textit{Hessian-Affine}~\cite{MiS04} keypoints are extracted from each image and are described by RootSIFT feature~\cite{AZ12}. Finally, the RootSIFT features from each image are pooled by VLAD~\cite{JPDSPS11} with a small visual vocabulary of size \textit{64}. The resulting \textit{8,192}-dimensional feature is further mapped  to \textit{512} dimensions by PCA. Following~\cite{JPDSPS11}, the final VLAD vector is normalized to unit length. In the direct k-way clustering case, we set the number of maximum iterations for all methods to \textit{20}. While for the bisecting case, there is no threshold on the number of iterations. The results reported in this section have been averaged over \textit{10} runs for each method.

\begin{figure}[t]
\begin{center}
	\subfigure[direct k-way]
    {\includegraphics[width=0.40\linewidth]{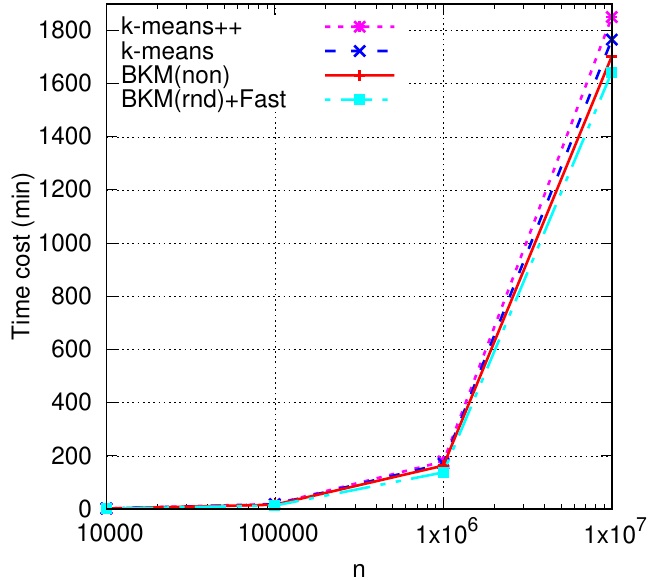}}    
	\subfigure[bisecting]
    {\includegraphics[width=0.38\linewidth]{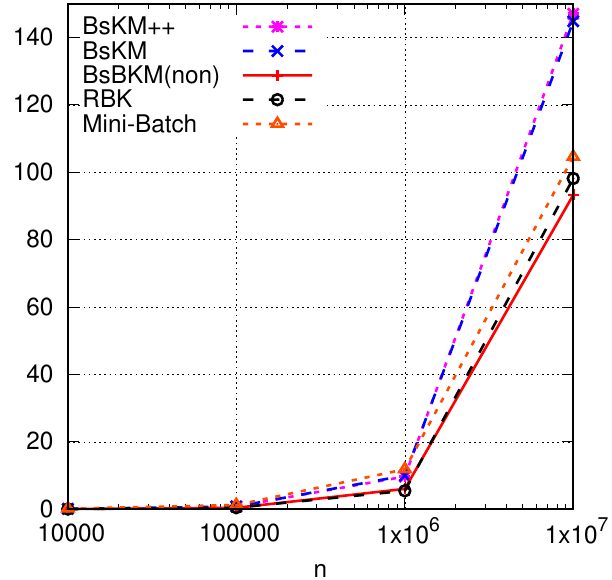}}    
	\subfigure[direct k-way]
    {\includegraphics[width=0.40\linewidth]{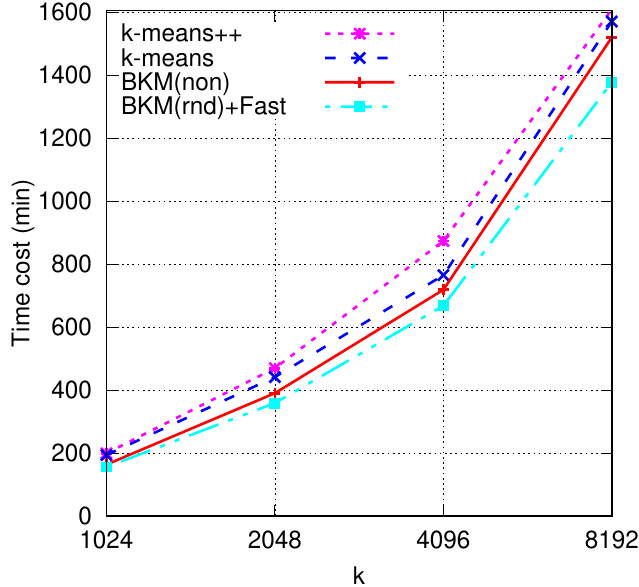}} 
	\subfigure[bisecting]
    {\includegraphics[width=0.38\linewidth]{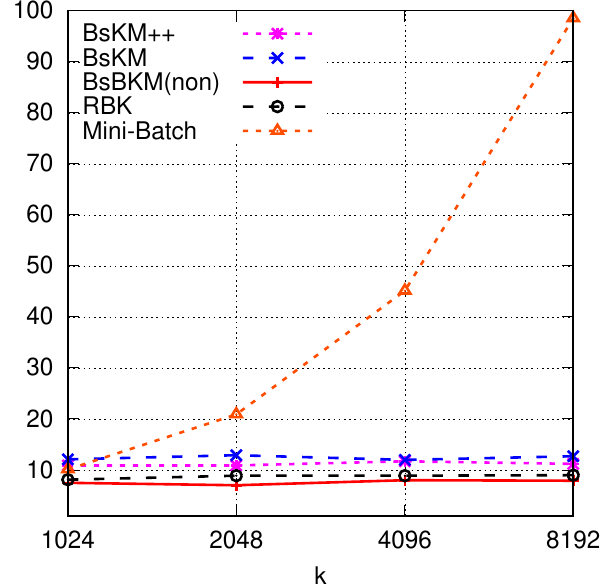}}   
	\caption{Scalability test by varying the scale of input data: (a)-(b) and by varying the number of clusters: (c)-(d). }
	\label{fig:tmkway}
\end{center}
\end{figure}

\begin{figure}
\begin{center}
	\subfigure[k=\textit{1024}, vary \textit{n}]
    {\includegraphics[width=0.48\linewidth]{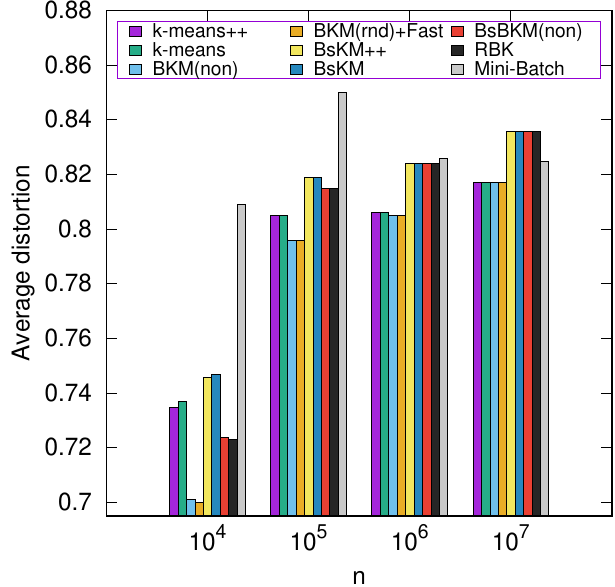}}
    \hspace{0.05in}
	\subfigure[n=$10^6$, vary \textit{k}]
    {\includegraphics[width=0.48\linewidth]{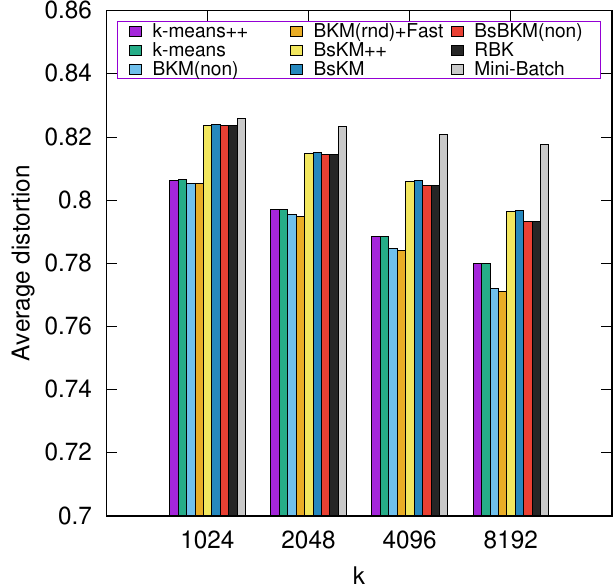}}    
	\caption{Average distortion from all \textit{9} methods under two different scalability testings on Flickr10M (best viewed in color).}
	\label{fig:dstall}
\end{center}
\vspace{-0.2in}
\end{figure}

In the first experiment, clustering methods are tested in the way that the scale of input images varies from 10K to 10M. A fixed number of clusters, i.e., \textit{1,024} is used regardless of the size of dataset. The time costs for direct k-way and bisecting methods are presented in Figure~\ref{fig:tmkway}(a)-(b). Accordingly, the average distortion of all the methods are presented in Figure~\ref{fig:dstall}(a).

As shown in the figures, BKM exhibits slightly faster speed over k-means and its variants across different scales of input data under both direct k-way and bisecting cases. The speed-up becomes more significant as the scale of input data increases. The higher efficiency of these methods is mainly attributed to no requirement of initial assignment. Compared with BKM(non), BKM(rnd)+Fast takes extra time. However, the cost of initial assigment is compensated later for not seeking the best movement. 
Compared with direct k-way clustering, methods with bisecting strategy achieve much higher scalability. In particular, BsBKM(non) shows the highest scalability. It only takes less than \textit{94} minutes to cluster \textit{10} million vectors (in \textit{512} dimensions) into \textit{1,024} clusters. The efficiency of Mini-Batch is close to BsBKM(non). However, as shown in Figure~\ref{fig:dstall}(a), the clustering quality is poor in most of the cases. Overall, BKM(rnd)+Fast achieves the highest speed efficiency and lowest distortion among all direct k-way clustering methods. While in the bisecting case, BsBKM(non) shows the best performance in terms of both speed efficiency and clustering quality. Similar to the experiments in Section~\ref{sec:distort}, the average distortion introduced by bisecting clustering is much higher than direct k-way due to the problem of under-fitting. 

In addition, the scalability of clustering methods is tested in the way that the number of clusters by varying from \textit{1,024} to \textit{8,192}, while the scale of input data is fixed to \textit{1} million. Figure~\ref{fig:tmkway}(c)-(d) show the time cost of all \textit{9} methods. Accordingly, the average distortion from all these \textit{9} methods are presented in Figure~\ref{fig:dstall}(b). As shown in the figures, for all direct k-way clustering methods, the time cost increases linearly as the number of clusters increases. Mini-Batch is no longer efficient as \textit{k} increases. In contrast, the time cost of all bisecting methods remains steady across different cluster numbers. In terms of clustering quality, as seen from Figure~\ref{fig:dstall}(b), in both direct k-way and bisecting cases, clustering driven by the proposed optimization procedure (Alg.~\ref{alg:kway}) performs considerably better. A clear trend is observed from Figure~\ref{fig:dstall}(b),  methods based on Alg.~\ref{alg:kway} shows increasingly higher performance than the rest as \textit{k} grows. Overall, clustering driven by proposed optimization process shows higher speed and better quality. The highest speed is achieved by BsBKM(non), for which only \textit{8} minutes are required to cluster \textit{1} million high dimensional data into \textit{8,192} clusters. Due to extra cost in initial assignment, bisecting with traditional k-means and k-means++ still shows around \textit{35\%} slower speed than BsBKM(non).

As a summary, clustering based on Alg.~\ref{alg:kway} shows superior performance in terms of both speed efficiency and quality under different scenarios. This is mainly due to the nature of incremental updating scheme, which allows the cluster structures to be fine-tuned in a more efficient way. When the proposed Alg.~\ref{alg:kway} is performed under bisecting manner (i.e., BsBKM(non)), it shows two orders of magnitude faster than traditional k-means.



\section{Conclusion}
\label{sec:conl}
We have presented a novel k-means variant. Firstly, a clustering objective function that is feasible for the whole $\textit{l}_2$-space is developed. Supported by the objective function, the traditional k-means clustering has been modified to simpler form. In this novel k-means variant, we interestingly find that neither the costly initial assignment nor the seeking of closest centroid for each sample in the iteration are necessary. This leads to higher speed and considerably lower clustering distortion. Furthermore when the proposed clustering method is undertaken in the ways of top-down bisecting, it achieves the highest scalability and best quality among all hierarchical k-means variants. Extensive experiments have been conducted in different contexts and on various datasets. Superior performance over most of the k-means variants is observed across different scenarios.

\section*{Acknowledgement}
This work is supported by National Natural Science Foundation of China under grants 61572408. The authors would like to express their sincere thanks to Prof. George Karypis from University of Minnesota, USA for his detailed explanation about the implementation of repeated bisecting k-means.

\section*{Appendix A: Optimality of Incremental K-means}
\label{sec:appa}
As shown in Eqn.~\ref{eqn:ii4} and Eqn.~\ref{eqn:iz3}, two optimal objectives are quite similar. In this section, we show that optimal solution with respect to objective function (Eqn.~\ref{eqn:iz3}) can be reached with incremental updating scheme presented in Alg.~\ref{alg:kway}.

\paragraph{Proof}: For contradiction, let $A^{o}=\{S_1,S_2,\cdots,S_k\}$ be an optimal solution and assume that there exists one element \textit{d} and clusters $S_i$ and $S_j$ such that $d \in S_i$. Now consider the clustering solution $A^*=\{S_1,S_2,\cdots,\{S_i-d\},\cdots,\{S_j+d\},\cdots,S_k\}$. Let $D_i$, $C_i$, and $D_j$, $C_j$ be the composite and centroid vectors of cluster $S_i-d$ and $S_j$, respectively. Let $e=\mathcal{I}_1(A^o)-	\mathcal{I}_1(A^*)$, then
\begin{equation}
\scriptsize{
\begin{aligned}
	e&=\frac{(D_i+d)'(D_i+d)}{n_i+1}+\frac{D_j'D_j}{n_j}-(\frac{D_i'D_i}{n_i}+\frac{(D_j+d)'(D_j+d)}{n_j+1}) \\
	&=(\frac{(D_i+d)'(D_i+d)}{n_i+1}-\frac{D_i'D_i}{n_i})-(\frac{(D_j+d)'(D_j+d)}{n_j+1} - \frac{D_j'D_j}{n_j}) \\
	&=\frac{2n_id'D_i+n_i{d'd}-D_i'D_i}{n_i(n_i+1)}-\frac{2n_jd'D_j+n_j{d'd}-D_j'D_j}{n_j(n_j+1)} \nonumber
\end{aligned}
}
\label{eqn:opt1}
\end{equation}
\normalsize{
Let's define $\mu_i=\frac{D_i'D_i}{n_i(n_i+1)}$, $\mu_j=\frac{D_j'D_j}{n_j(n_j+1)}$ are the average pairwise inner product in cluster $S_i$ and $S_j$ respectively. In addition, $\delta_i$ and $\delta_j$ are given as the average inner-products between \textit{d} and elements in $S_i$ and $S_j$ respectively, viz $\delta_i=\frac{d'D_i}{n_i}$, and $\delta_j=\frac{d'D_j}{n_j}$. Above Equation is rewritten as}
\begin{equation}
\scriptsize{
\begin{aligned}
	e&=(\frac{2n_i\delta_i}{n_i+1}+\frac{d'd}{n_i+1}-\frac{n_i\mu_i}{n_i+1}) - (\frac{2n_j\delta_j}{n_j+1}+\frac{d'd}{n_j+1}-\frac{n_j\mu_j}{n_j+1}) \\
	&\approx (2\delta_i-2\delta_j+\frac{d'd}{n_i+1})-(\mu_i-\mu_j+\frac{d'd}{n_j+1})
\end{aligned}
}
\label{eqn:opt2}
\end{equation}
\normalsize{
Given the fact that $(2\delta_i-2\delta_j+\frac{d'd}{n_i+1}) < (\mu_i-\mu_j+\frac{d'd}{n_j+1})$, we have $\mathcal{I}_1(A^o) <	\mathcal{I}_1(A^*)$, which is contradicting. $\blacksquare$

\section*{Appendix B: Convergence of Incremental k-means}
\label{sec:appb}
\normalsize{
$S_i$ and $S_j$ are two clusters. \textit{d} is initially part of $S_i$, and $D_i$ is the composite of $S_i$ exclude \textit{d}, $C_i$ is the centroid of $S_i$ exclude \textit{d}, $D_j,C_j$ is the composite and centroid of cluster $S_j$, the move condition of \textit{d} from $S_i$ to $S_j$ should satisfied}
\begin{equation}
\scriptsize{
\begin{aligned}
	\frac{(D_i+d)'(D_i+d)}{n_i+1}+\frac{D_j'D_j}{n_j}<\frac{D_i'D_i}{n_i}+\frac{(D_j+d)'(D_j+d)}{n_j+1} 
\end{aligned}
}
\label{eqn:app1}
\end{equation}
\normalsize{
This equation can be rewritten as:}
\begin{equation}
\scriptsize{
\begin{aligned}
	\frac{(D_i+d)'(D_i+d)}{n_i+1}-\frac{D_i'D_i}{n_i} &< \frac{(D_j+d)'(D_j+d)}{n_j+1}-\frac{D_j'D_j}{n_j} \\
	\frac{D_i'D_i+2d'D_i+d^2}{n_i+1}-\frac{D_i'D_i}{n_i} &<  \frac{D_j'D_j+2d'D_j+d^2}{n_j+1}-\frac{D_j'D_j}{n_j}\\
	\frac{2n_id'D_i+n_id^2-D_i'D_i}{n_i(n_i+1)} &<  \frac{2n_jd'D_j+n_jd^2-D_j'D_j}{n_j(n_j+1)} \\
	2\frac{n_i}{n_i+1}\frac{d'D_i}{n_i}-\frac{D_i'D_i}{n_i(n_i+1)}+\frac{d^2}{n_i+1} &< 2\frac{n_j}{n_j+1}\frac{d'D_j}{n_j} \\-\frac{D_j'D_j}{n_j(n_j+1)}+\frac{d^2}{n_j+1} \nonumber
\end{aligned}
}
\end{equation}
\normalsize{
Now if we assume that both $n_i$ and $n_j$ are sufficiently large, then $\frac{n_i}{n_i+1}$ and $\frac{n_j}{n_j+1}$ will be close to \textit{1}. Under these assumptions, we can get }
\begin{equation}
\scriptsize{
\begin{aligned}
2\frac{d'D_i}{n_i}-\frac{D_i'D_i}{n_i(n_i+1)}+\frac{d^2}{n_i+1} < 2\frac{d'D_j}{n_j}-\frac{D_j'D_j}{n_j(n_j+1)}+\frac{d^2}{n_j+1}. \nonumber
\end{aligned}
}
\end{equation}
\normalsize
Now $\mu_i=\frac{D_i'D_i}{n_i(n_i+1)}$, $\mu_j=\frac{D_j'D_j}{n_j(n_j+1)}$ are defined as the average pairwise inner product in cluster $S_i$ and $S_j$ respectively. $\delta_i$ and $\delta_j$ are given as the average inner-products between \textit{d} and elements in $S_i$ and $S_j$ respectively, viz $\delta_i=\frac{d'D_i}{n_i}$, and $\delta_j=\frac{d'D_j}{n_j}$, the following inequation holds. 
 \begin{equation}
 \scriptsize{
 2\delta_i-2\delta_j+\frac{d'd}{n_i+1} < \mu_i-\mu_j+\frac{d'd}{n_j+1}.
 }
 \end{equation}

\bibliographystyle{ieeetr}  
\bibliography{wlzhao}

\end{document}